\documentclass[utf8]{FrontiersinHarvard} 

\usepackage{url,hyperref,lineno,microtype,subcaption}
\usepackage[onehalfspacing]{setspace}

\usepackage[whole]{bxcjkjatype} 

\usepackage{amsmath,amsthm,ascmac}

\usepackage{algorithm}
\usepackage{algpseudocode}

\theoremstyle{plain}

\newtheorem*{thm*}{Theorem}
\theoremstyle{remark} 

\usepackage{diagbox}
\usepackage{enumerate}
\usepackage{multirow}
\usepackage{multibib}
\usepackage[normalem]{ulem}

\usepackage[final,protect]{changes} 
\setaddedmarkup{\textcolor{red}{#1}}

\theoremstyle{plain}

\theoremstyle{definition}

\theoremstyle{remark}

\def\keyFont{\fontsize{8}{11}\helveticabold }
\def\firstAuthorLast{
} 
\def\Authors{
Tadahiro Taniguchi\,$^{1,2*}$, Masafumi Oizumi\,$^{3}$ , Noburo Saji\,$^{4}$, Takato Horii\,$^{5}$  and Naotsugu Tsuchiya\,$^{6}$}
\def\Address{
$^{1}$ Graduate School of Informatics, Kyoto University, Kyoto, Japan\\
$^{2}$ Research Organization of Science and Technology, Ritsumeikan University, Kyoto, Japan\\
$^{3}$Graduate School of Arts and Sciences, The University of Tokyo, Tokyo, Japan \\
$^{4}$Faculty of Human Science, Waseda University, Tokorozawa, Japan \\
$^{5}$Graduate School of Engineering Science, Osaka University, Osaka \\ 
$^{6}$School of Psychological Sciences, Monash University, Melbourne, Australia \\ 
$^{7}$ Center for Information and Neural Networks (CiNet), National Institute of Information and Communications Technology (NICT), Suita-shi, Osaka
, Japan \\
$^{8}$ 
Laboratory of Qualia Structure, ATR Computational Neuroscience Laboratories,
Soraku-gun, Kyoto
, Japan.
}
\def\corrAuthor{
Tadahiro Taniguchi}

\def\corrEmail{
taniguchi@i.kyoto-u.ac.jp}

\begin{document}
\onecolumn
\firstpage{1}

\title{Constructive Approach to Bidirectional \deleted{Causation}\added{Influence} between Qualia Structure and Language Emergence} 

\author[\firstAuthorLast ]{\Authors} 
\address{} 
\correspondance{} 

\extraAuth{}

\maketitle

\begin{abstract}
\added{This perspective paper explores}\deleted{This paper presents a novel perspective on } the bidirectional \deleted{causation}\added{influence} between language emergence and the relational structure of subjective experiences, termed qualia structure, and lays out a constructive approach to the intricate dependency between the two.
\deleted{We hypothesize that languages with distributional semantics, e.g., syntactic-semantic structures, may have emerged through the process of
aligning internal representations among individuals, and such alignment of internal representations facilitates more structured language. This mutual dependency is suggested by the recent advancements in AI and symbol emergence robotics, and in collective predictive coding (CPC) hypothesis, in particular.}
\added{We hypothesize that the emergence of languages with distributional semantics (e.g., syntactic-semantic structures) is linked to the coordination of internal representations shaped by experience, potentially facilitating more structured language through reciprocal influence. This hypothesized mutual dependency connects to recent advancements in AI and symbol emergence robotics, and is explored within this paper through theoretical frameworks such as the collective predictive coding.}
Computational studies show that neural network-based language models form systematically structured internal representations, and multimodal language models can share representations between language and perceptual information. This perspective suggests that language emergence serves not only as a mechanism creating a communication tool but also as a mechanism for allowing people to realize shared understanding of qualitative experiences. The paper discusses the implications of this bidirectional \deleted{causation}\added{influence} in the context of consciousness studies, linguistics, and cognitive science, and outlines future constructive research directions to further explore this dynamic relationship between language emergence and qualia structure.
\tiny
 \keyFont{ \section{Keywords:} qualia structure, symbol emergence, large language model, representation learning, emergent systems, consciousness} 
\end{abstract}

\section{Introduction}
The relationship between language and qualia---qualitative aspects of consciousness---is an intriguing research topic~\citep{WinawerRussiancolor,zlatev2008dialectics}. One of the closest contacts between the two fields has been the issue of the dependency of perception on language as studied in developmental psychology and cross-linguistic studies. \deleted{For example, the Sapir-Whorf hypothesis has been tested in the context of developmental psychology and cross-linguistic studies, connecting the two fields.}\added{For example, the Sapir-Whorf hypothesis \citep{deutscher2010through}—which posits that the structure of a language can influence its speakers’ patterns of thought and perception—has been tested in these contexts, providing a conceptual bridge between the two domains.}

This paper considers a related, yet distinct issue: bidirectional \deleted{causation}\added{influence} between language emergence and qualia structure ~\citep{kawakita2025my,tsuchiya2024qualia,kleiner2024towards, Tsuchiya2021-yp, Lyre2022-ht, Tallon-Baudry2022-yn, Malach2021-zn, Fink2021-jy}.
\deleted{A qualia structure is the structure of relationships between different qualia, such as the quale of red being closer to pink than to blue.}
\added{Relationships between different qualia, such as the quale of red being closer to pink than to blue, define what we term a qualia structure. This notion has roots in the classic inverted‑spectrum debate, where structural similarity within a quality space was taken to underpin phenomenal character \citep{shoemaker1982inverted, white1985shoemaker}. Recent computational work—including our own—builds on this idea by quantifying such structural relations with data‑driven metrics~\citep{Kawakita2023-xe}.}
\deleted{This} \added{Computational focus in this paper} is mainly motivated by findings in artificial intelligence (AI) and symbol emergence in robotics~\citep{Mikolov2013,Huh2024-ax,taniguchi2016symbol,taniguchi2019symbol}. Neural network-based language models are known to form internal representations that hold structural relationships. Also, multimodal language models form internal representations shared by language and perceptual information. Artificial cognitive systems develop internal representations\footnote{In this paper, ``internal representations'' simply refer to neural activation patterns in natural and artificial neural networks. Note that the usage of the term ``internal representations'' does not imply that the authors hold computationalism as in classical symbolic AI~{\citep{Newell1980}}.} through a dual process: bottom-up integration of sensory-motor information and top-down incorporation of linguistic knowledge, resulting in multimodal internal representations. Language, by imposing structure on these representations, may significantly influence how sensory-motor information is organized internally. For humans, sharing language may lead to structurally similar qualia structures across individuals.  Thus, language serves not just as a communication tool, but as an aligning force for subjective experiences.  Harmonization of sensory-motor and linguistic information into internal representations fundamentally shapes our perception, interpretation, and interaction with the world, underscoring the profound influence of language on human cognition and conscious experience, while language itself emerges in human society on the basis of our physical and social interactions, i.e., language emergence.
Although the exact relationship between qualia and neural activity in the brain (i.e., internal representations) has not yet been elucidated, in this paper we assume that the relational structure of neural activity would correlate with the relational structure of qualia. \deleted{Thus, we assume that a change in the structure of internal representations leads to a change in the structure of qualia.} \added{For the specific purpose of exploring language-structure interactions via the constructive approach outlined here, we proceed under the working assumption that modifications to the relational structure of internal representations, as manipulated or observed in our models, can serve as indicators of potential modifications to the corresponding qualia structure. We acknowledge, however, that the precise mapping between representational structure and qualia structure is known to be complex and likely indirect.}\footnote{\added{Indeed, the actual relationship between the structure of internal representations and qualia structure is highly complex, rather than a simple one-to-one mapping assumed here as a working hypothesis. Evidence from various domains, including recovery processes in neurorehabilitation \citep{stefaniak2011reorganization}, phenomena like anosognosia for visual deficits \citep{michel2024when}, and studies on the neural correlates of perception \citep{pollen1999neural}, suggests that significant changes in neural states or representations do not always entail straightforward changes in subjective experience. Such complexities may potentially be understood through concepts like neural plasticity and degeneracy (i.e., the ability of structurally different neural systems to yield the same output or function) \citep{tononi1998consciousness, price2002degeneracy}. A detailed analysis of this intricate neural-phenomenal mapping problem is beyond the scope of the present paper, which focuses specifically on the interplay between emergent language and qualia structure.}}

\added{We use qualia in the broad sense proposed by \citep{chalmers2002components}: the subjective ``what‑it‑is‑like'' character of conscious states—for example, what it is like to see green or to feel a toothache. This definition carries no commitment to claims about qualia being intrinsically private or ineffable. Because our goal is to analyse how such experiential qualities are organized rather than to address the metaphysical ``hard problem,'' we retain the concise term qualia instead of longer equivalents such as quality of experience. We refer to the relational pattern formed by these experiential qualities as a qualia structure. This notion corresponds to what others call a quality‑space structure, but we adopt ``structure'' because it is both briefer and, mathematically, more general, allowing for richer relational descriptions beyond simple metric spaces (\cite{kleiner2024towards}; see also \cite{lee2023structuralism_psyarxiv}). While the present paper does not attempt to resolve the hard problem, we assume—following \cite{kawakita2025my}—that certain phenomenal properties can be empirically investigated by focusing on the relationships among qualia.}

In psychological studies, language is considered to exert some top-down influence on perceptual internal representations. It is natural to assume that speakers of the same language will share a similar structure of internal representations to some degree, despite the private nature of their subjective perceptual experiences (i.e., the fact that one cannot directly access or experience another person's sensations, thoughts, or qualia).
However, notably, as a famous philosophical riddle of ``inverted qualia'' implies, structural relationships between language and qualia are not straightforward. Given a set of the same external stimuli, qualia for the corresponding stimuli can be consistently labeled and communicated with each other even if qualia are inverted between two persons. Thus, structural equivalence at the level of language does not guarantee that of qualia structure. \deleted{One possible approach to the qualia inversion problem is to evaluate a structural similarity between individuals in an unsupervised, i.e., language-free manner, and to consider the unsupervised structural similarity as one of the necessary conditions to guarantee no qualia inversion~citep{kawakita2025my}.}

Meanwhile, linguistic communication faces similar challenges in sharing meanings. Consider that the meanings of two words are swapped between the two persons, yet they do not notice it. This is a rather likely scenario. \deleted{Yet people manage to understand each other through language thanks to the relations between words and their associated qualia, i.e., subjective (multimodal) experiences.}
\added{While the mechanisms underlying mutual understanding through language are complex and debated (involving factors such as triangulation, which relies on joint attention towards external referents, and pragmatic inference \citep[e.g.,][]{davidson1984inquiries, grice1989studies}), there is also an intuition that some alignment occurs at the level of internal representations linked to subjective experience. This paper does not offer a complete theory of understanding, but instead focuses on exploring this potential alignment, specifically how the structure of emergent language might influence the structure of internal representations. We suggest that a constructive approach, using data-driven computational models (like the collective predictive coding framework and Metropolis-Hastings naming games discussed later, which explicitly incorporate joint attention), provides a promising alternative to purely speculative methods for investigating these underlying alignment mechanisms.}

This \added{challenge of grounding shared meaning} is related to the symbol grounding problem in cognitive science and AI~\citep{harnad1990symbol}.
This implies that there are underlying mechanisms in our brains, the world, and society that enable the emergence of communication and symbol systems that share \deleted{structurally similar}\added{coordinated}  internal representations \added{and sign usage}.
\added{Note that ``emergence'' in the context of language or symbol emergence, as discussed here, refers to weak emergence, where higher-level patterns arise from lower-level interactions but supervenience is maintained (e.g., \cite{bedau1997weak}). Research in this field investigates both how linguistic properties (such as structure, usage, and meaning) form through interactions among agents and, reciprocally, how these emergent systems subsequently influence agents' perception and behavior (e.g., \cite{Steels2005,taniguchi2016symbol}).}
Investigating the mutual dependency between language emergence and qualia structure alignment, therefore, may shed an interesting light to address an age-old problem in the study of consciousness.

To study the formation and alignment of qualia structure and its dependency to language, we \deleted{argue}\added{think} that taking constructive approaches is promising.
The constructive approach, which is often employed in complex systems, neuroscience, and cognitive science, is a methodology that focuses on building and analyzing artificial systems to understand how the target system works~\citep{pfeifer2001understanding,asada2001cognitive,taniguchi2016symbol}. 
By building artificial systems that simulate cognitive processes, we can observe how internal representations emerge and evolve. These representations can then be analyzed in the \deleted{constructive} model.
Critically, the constructive approach allows us to explore the potential causal mechanisms behind the formation of qualia structures in a controlled, observable environment. It provides a way to bridge the gap between abstract philosophical concepts and concrete, implementable models, potentially offering new insights into the nature of subjective experience and its structural properties.

As such, this paper introduces a constructive approach and explains how this approach can address the issue of bidirectional \deleted{causation}\added{influence} between qualia structure formation and language emergence (see Section 2).
Bidirectional \deleted{causation}\added{influence} in the context of emergent and complex systems refers to the idea that causality in such systems is not unidirectional, but instead involves mutual influences between different levels or components of the system~\citep{kalantari2020emergence,Polanyi1966,maturana1980autopoiesis}. This means that while lower-level components (e.g., individual agents or parts of the system) can form higher-level patterns or orders to emerge (\deleted{upward causation}\added{upward organization}), the emergent higher-level properties or structures can, in turn, influence or constrain the behavior of the lower-level components (downward \deleted{causation}\added{constraint}).

In studies on language emergence and symbol emergence in robotics, it is considered that language is formed in society on the basis of human perceptual experiences in a bottom-up manner~\citep{Steels2005,lazaridou2020emergent,taniguchi2016symbol,taniguchi2019symbol}. At the same time, the formed language gives top-down constraints to people's perception and actions. This kind of bidirectional \deleted{causation}\added{influence} is introduced in the holistic view of language emergence, called symbol emergence systems~\citep{taniguchi2016symbol,taniguchi2019symbol}.
With this view, a series of constructive studies \deleted{called}\added{on} symbol emergence in robotics have been conducted. Recently, \citet{taniguchi2024collective} proposed an idea called collective predictive coding (CPC) and provided a computational and theoretical relationship between cognitive systems modeled by predictive coding and the free-energy principle, and symbol or language emergence, explicitly modeling the bidirectional \deleted{causation}\added{influence} in language emergence mathematically. Extending this view, this paper is the first to discuss the intersection of bidirectional \deleted{causation}\added{influence} in language emergence and qualia structure.

\begin{figure}[t]
    \centering
    \includegraphics[width=0.6\linewidth]{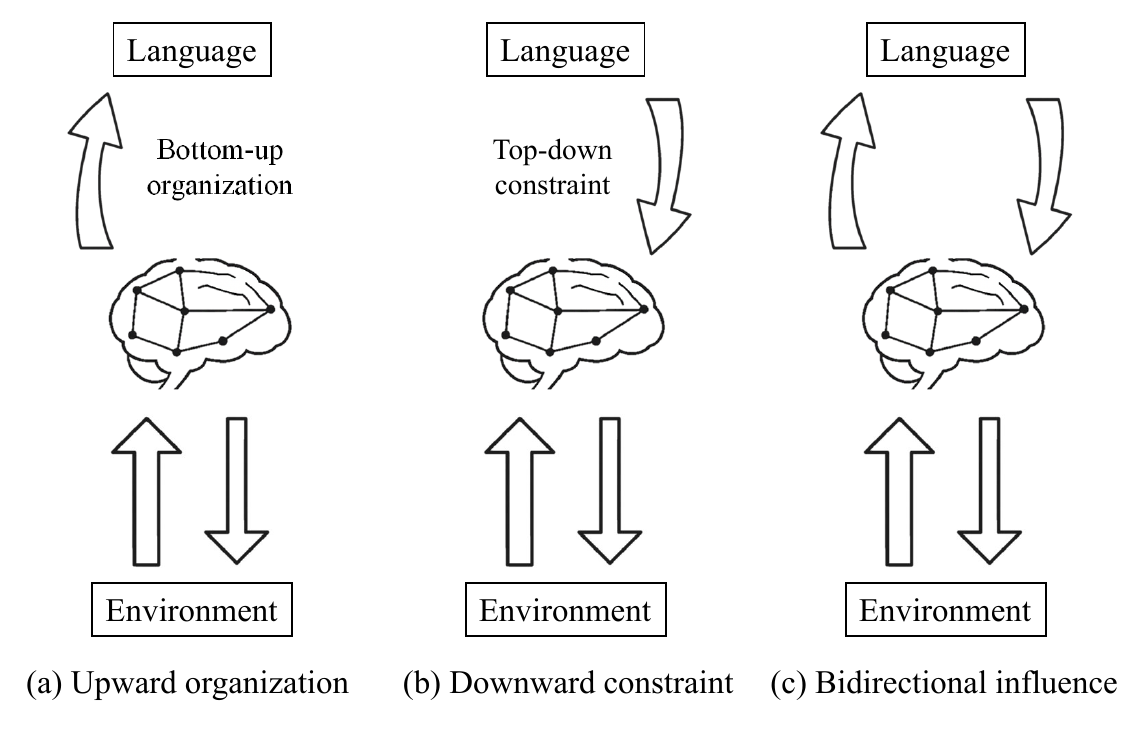}
    \caption{\deleted{(a) Emergence of language from internal representations.
(b) Effect of language on internal representations.
(c) Mutual influence between language and internal representations. \\Note that the existence of multiple agents (i.e., multiple brains) is crucial for the bottom-up organization of language, although we describe a single brain for each panel to avoid complicating the figure.}\added{Schematic diagrams illustrating bidirectional influence between language and internal representational structure. (a) Upward organization: Internal representational structure, formed through perceptual experience of the environment, influences language emergence. (b) Downward constraint: Language structure affects the structure of internal representations. (c) Bidirectional influence: Mutual interaction combining upward organization and downward constraint. Note: Each brain depicted internally forms and possesses a representational structure (illustrated schematically). The existence of multiple agents (i.e., multiple brains) is crucial for the actual bottom-up organization of language, although we describe a single brain for each panel to avoid complicating the figure.}}
    \label{fig:lang-struct}
\end{figure}

The rest of the paper is organized as follows. Section 2 discusses the emergence of language and qualia structure, introducing the concept of bidirectional \deleted{causation}\added{influence} and CPC. It also explores psychological perspectives on language and perception, and introduces a hierarchy of qualia levels. Section 3 explores the constructive approach to the mutual influence between qualia structure and language emergence, examining representational alignment and structural influences both within individuals and between people. It also discusses the constructive approach to investigating this relationship, including recent developments in neural network models and representational similarity analysis. Section 4 presents a discussion and concludes this paper.

\section{Emergence of language and qualia structure}

\subsection{Bidirectional \deleted{causation}\added{influence} between qualia and language}

Qualia affects language and vice versa.
Considering the phenomena of language (or symbol) emergence, (internal) representation learning, and (grounded) language acquisition, this \deleted{mutual influence }can be regarded as bidirectional \deleted{causation}\added{influence}.
This section summarizes the perspectives on the bidirectional \deleted{causation}\added{influence} between qualia and language following the diagram shown in Figure~\ref{fig:lang-struct}.

First, Figure~\ref{fig:lang-struct} (a) shows the \deleted{upward causation}\added{upward organization}. We should not forget that language systems evolve on top of our brains through human biological and cultural adaptation to the environment~\citep{deacon1998symbolic,Christiansen_Morten_H_2023-05-04,kirby2002emergence,Steels2005}. We employ language as a means of adapting to our environment, enabling us to engage in the physical and social interactions necessary for thriving in our environment. From a developmental perspective, Piaget conventionally introduced the idea of a schema system that is structurally formed through physical interactions with an infant's environment using sensory-motor systems~\citep{Flavell}. \deleted{He described a dynamic view of self-organization of cognitive systems.}\added{He described how cognitive systems dynamically develop their structure through interaction.} In the context of AI and cognitive robotics, this process is represented by representation learning (or world model learning~\citep{friston2021world}). On such a basis, language emerges within a society based on cognitive systems~\citep{taniguchi2019symbol,taniguchi2024collective}. 
Through representation learning, agents learn internal representations that model the world's structure from multimodal perceptual information. As a result, the structure of these learned internal representations may come to reflect the structure of the world to a certain extent~\footnote{Note that, as qualia are subjective experiences, their structure does not necessarily reflect the structure of the objective world. Qualia structures can accommodate phenomena such as dreams, hallucinations, and other non-veridical experiences. For a more detailed discussion on this topic, please see \cite{prakash2020fact}.}.

Considering that language emerges on the basis of internal representations reflecting individuals' experienced worlds, the structure of language, which is encoded as distributional semantics, may collectively reflect \deleted{the world's structure}\added{the structure inherent in the internal representations formed by the participating individuals through their diverse experiences} as well~\citep{Huh2024-ax,taniguchi2024collective,taniguchi2024generative}. This is the \deleted{upward causation}\added{upward organization} between qualia and language. This topic is related to the emergence of language and symbol systems. Note that language (or symbol) emergence is not just in a single brain, but is \deleted{performed}\added{achieved} through the interaction between many brains via \deleted{semiotic interactions}\added{communication using language} (see Figure 2 and the caption of Figure 1).

Second, Figure~\ref{fig:lang-struct} (b) shows the \deleted{downward causation}\added{downward constraint}. This has been a long-standing topic regarding the effect of language on thoughts and perceptions (see Section 2.2).
\deleted{Recent evidence obtained through a constructive approach}\added{Evidence from the field of natural language processing} shows that distributional semantics can reconstruct structural relationships between concepts (see Section 3.3) 
(e.g., ~\cite{Mikolov2013}). \added{There is a substantial body of psycholinguistic research investigating the influence of language on perception. For example,~\cite{WinawerRussiancolor} demonstrated that Russian speakers, who lexically distinguish between light blue and dark blue, were faster at performing perceptual discrimination tasks involving these two shades than English speakers, whose language does not make this lexical distinction. Furthermore, ~\citep{ThierryColorPNAS} employed event-related potentials (ERPs) to examine the temporal dynamics of such language-related perceptual differences following stimulus presentation. Their findings revealed that these effects emerged as early as 100 ms post-stimulus, a time window associated with activity in the primary and secondary visual cortices. These results suggest that early stages of color perception can be modulated by one's native language. Furthermore, recent studies have argued that since closely related non-human primates do not exhibit human-like color categories, the emergence of consensus-based color categories in humans likely depends on language (e.g., ~\cite{Daniel2025colororigin}).} \added{These findings suggest that language play a significant role in shaping human perceptual categorization. Furthermore,} distributional semantics argues that a word's meaning can be determined by analyzing its statistical distribution and co-occurrence patterns with other words in large text corpora. Large language models (LLMs) exploit the nature of distributional semantics. \deleted{However, to what extent the structure of language affects qualia structure in human consciousness is still a mystery.}
\added{However, while distributional semantics derived from linguistic co-occurrence is well-studied in NLP, how this language structure interacts with representations derived from perceptual experience, and how this interplay influences the overall representational structure relevant to qualia, remains insufficiently explored –- particularly from a constructive modeling perspective aimed at understanding the formation of qualia structure.}

Third, Figure \ref{fig:lang-struct} (c) shows the bidirectional \deleted{causation}\added{influence}\deleted{, which has not yet been investigated in the literature}. The bidirectional \deleted{causation}\added{influence} is the combination of upward and \deleted{downward causation}\added{downward constraint}. 
 This kind of dynamic systems view of language and internal representation has been referred to as {\it  symbol emergence systems}~\citep{taniguchi2016symbol,taniguchi2019symbol}.
Even if one argues that language has a dominant effect on internal representations, language itself emerges in human society based on multimodal sensory-motor perceptual experiences and social interactions. This is a discussion of language and symbol emergence on embodied cognitive systems. 
Therefore, an important question to ask regarding the bidirectional \deleted{causation}\added{influence} is how humans could form a language that involves a structure capable of exerting such influence on qualia structure.

If we stand at the viewpoint of bidirectional \deleted{causation}\added{influence}, we consider that both qualia structure and language are \deleted{emergent or self-organized}\added{dynamically evolving}. \deleted{They are changing in a cyclic manner.}\added{They change through ongoing reciprocal influence.} So, what is the source of information for organizing structures of language and qualia? 
We hypothesize that one of the essential sources for both language and qualia structures is our embodied multimodal experience of the world.

\subsection{Psychological perspectives}
 How has psychology addressed the issue of bidirectional \deleted{causation}\added{influence} between language and perception? Specifically, what is known about the causal effects of language on perception, and what role does perceptual experience play in shaping the structure of language?
 
 The first question, often associated with the so-called Sapir-Whorf hypothesis, has attracted significant attention from psychologists, linguists, and anthropologists. Empirical studies have approached this issue with  two main questions in mind: how speakers of different languages perceive the world (e.g.,~\cite{berlin1969basic}) and how children's perception of the world changes as they acquire their native languages (e.g.,~\cite{Robersondev2004}). Research to date has revealed two key findings: 1) biologically based perceptual experience forms the foundation for language acquisition, and 2) individual languages exert a powerful influence on the perceptual experiences of speakers. A well-known traditional example of the first is the study in the domain of color. Linguistic and anthropological studies have reported that many languages share a  part of color lexicon, suggesting the existence of a universal perceptual basis in human beings (\cite{berlin1969basic, Regierfocalcolorsuniversal2005}). Moreover, developmental studies have demonstrated that even pre-linguistic infants perceive such colors categorically (\cite{yang2016cortical}). \added{\cite{Morigucyhicolor2025} examined how children and adults in Japan and China judge the relationships among multiple colors using a similarity judgment task. Remarkably, the results revealed no substantial differences in the structure of color similarity across cultures or age groups. These findings, we believe, further suggest the existence of universal aspects of color categorization.} The second point has sparked more intense debate. While many earlier studies reported that the influence of language on perception is limited to experimental tasks where language can be implicitly engaged (e.g., \cite{WinawerRussiancolor, roberson2000color, davidoff1999colour, Kay2006color} ), more recent studies increasingly show that language effects manifest even in purely non-linguistic contexts (e.g.,\cite{ThierryColorPNAS}). \added{Thus, the current focus of psychological research is not whether language affects perception (i.e., the issue originally asked in the Sapir-Whorf hypothesis), but rather in what task contexts and to what extent language influences perception \citep{imai2016relation}.} However, current evidence, \deleted{considering}\added{considered} across all cognitive and perceptual domains, has not reached convergence. It is possible that the degrees of these influences are highly dependent on various factors that are currently unknown.

The second question---how perceptual experience influences language structure---has been primarily explored by linguists. Cognitive linguists, in particular, have long examined how the structure of language is \deleted{motivated}\added{influenced} by perceptual experience, aiming to uncover how the abstract meanings and grammatical structures are grounded in our bodies (e.g., \cite{lakoff1987woman}). However, most researchers in this area employ descriptive approach, making it challenging to provide empirical evidence on this motivational issue. Especially the descriptive approach cannot establish a causal link between factors, which a constructive approach can complement.
Thus, a new approach is needed to address the relationships between language and perception, whether considering upward or downward \deleted{causation}\added{influence}.

\subsection{Collective Predictive Coding and the Bidirectional \deleted{Causation}\added{Influence}}

\begin{figure}
    \centering
    \includegraphics[width=0.9\linewidth]{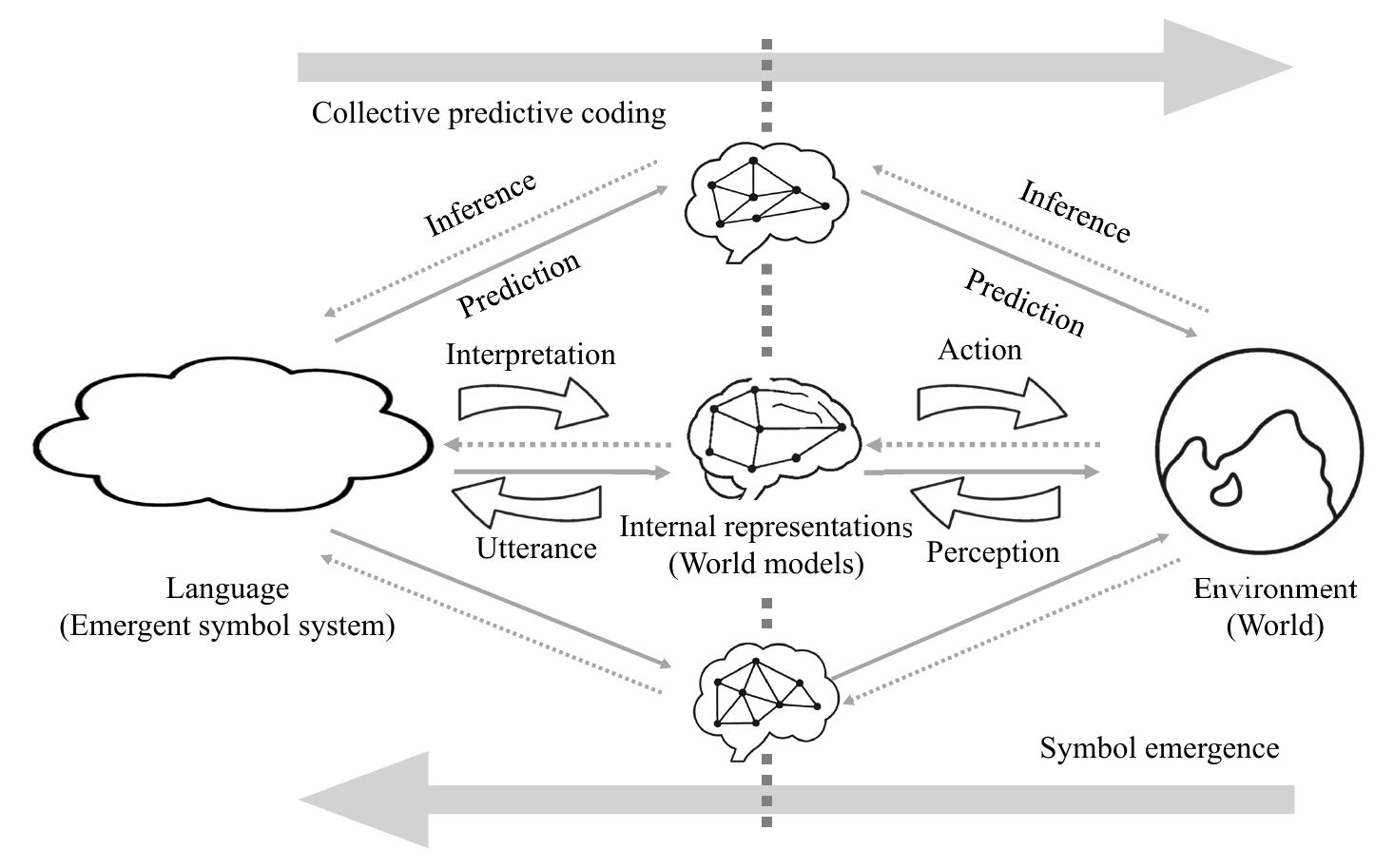}
    \caption{Shematic diagram of collective predictive coding.}
    \label{fig:birateral-cpc}
    \includegraphics[width=0.6\linewidth]{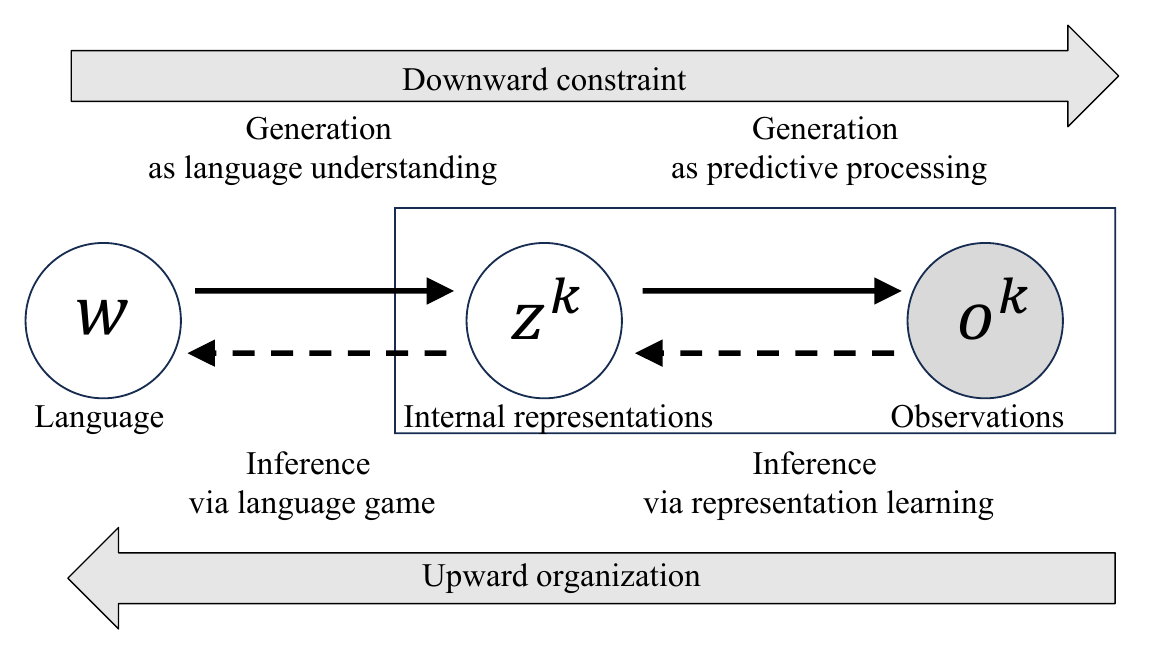}
    \caption{Probabilistic graphical model representing the CPC and bidirectional \deleted{causation}\added{influence} involved in symbol emergence systems.}
    \label{fig:pgm}
\end{figure}

Bidirectional \deleted{causation}\added{influence} is, generally, a characteristic feature of emergent systems\footnote{Emergent systems are complex systems that have emergent properties. Orders or structures that emerge in a bottom-up manner at the higher hierarchical level give top-down constraints onto the agents who interact with each other. This bidirectional \deleted{causation}\added{influence} brings about emergent functions to the system.}.
\cite{taniguchi2016symbol,taniguchi2019symbol} \added{put forward the view}\deleted{argued} that symbolic communication, including language, is an emergent function that is based on this type of emergent system. Specifically, they call the system a {\it symbol emergence system}. 
The collective predictive coding (CPC) hypothesis \deleted{argues}\added{posits} that language emerges through CPC~\citep{taniguchi2024collective}.
Figures~\ref{fig:birateral-cpc} \added{and ~\ref{fig:pgm}} show the diagram and probabilistic graphical model (PGM) representation of the CPC\added{, respectively}. Throughout the paper, in the CPC, $w$, $z$ and $o$ represent language, internal representations, and (perceptual) observations, respectively, following the convention of the original and related papers.
This idea extends the concept of predictive coding~\citep{hohwy2013predictive} from the individual cognitive system level to the collective (social) systems level, and provides a viewpoint to formalize the symbol (language) emergence as a decentralized Bayesian inference~\citep{hagiwara2019symbol,taniguchi2023emergent}. In other words, language as a distribution of sentences is sampled from the posterior distribution of latent variables given agents' distributed observations.
This idea is closely related to the free-energy principle and world models~\citep{friston2019free,Parr_Thomas2022-03-29,friston2021world,taniguchi2023world}.

\deleted{The main argument of this paper is to extend the idea to the structural level.} \added{This paper extends the idea of CPC to the structural level and lays out future directions for the constructive study of qualia structure.} Specifically, we aim to clarify the mutual influence between the latent syntactic-semantic structure of language, i.e., distributional semantics, and the structure of perceptual internal representations (Figure~\ref{fig:influence}). Additionally, we \deleted{argue}\added{suggest} that the structures of internal representations of agents are influenced and becoming similar when mediated by language (Figure~\ref{fig:influence}).

The CPC in the simplest case can be described mathematically as follows:
\begin{align}
\text{Generative model:}\quad & p\left(\{o^k\}_k, \{z^k\}_k, w\right)=p(w) \prod_k p\left(o^k \mid z^k\right) p\left(z^k \mid w\right)\ \label{eq:generate}\\
\text{Inference model:}\quad & q\left(w, \{z^k\}_k \mid \{o^k\}_k\right) = q\left(w \mid \{z^k\}_k\right) \prod_k q\left(z^k \mid o^k\right) \label{eq:infer}
\end{align}

\added{Here, $o^k$ represents the perceptual observation of the $k$-th agent, $z^k$ is their corresponding internal representation (a latent variable), and $w$ is a latent variable representing the shared language (e.g., a word or sentence). The notation $\{ \cdot \}_k$ denotes the set of variables for all agents $k$ in the group.}

\added{Equation~\ref{eq:generate} describes the assumed generative process: it posits that the shared language ($w$) influences each agent's internal representation ($z^k$) (cf. downward constraint), which in turn is used to predict that agent's perceptual observation ($o^k$). The term $p(w)$ represents the prior probability of the language itself.}

\added{Equation~\ref{eq:infer} describes the corresponding inference process (often approximated using variational inference, hence the notation $q(\cdot)$ instead of $p(\cdot)$): it models how agents might estimate the shared language ($w$) and their own internal representations ($z^k$) given their observations ($\{o^k\}_k$). This inference is decomposed into two parts: first, each agent $k$ infers their own internal representation $z^k$ from their observation $o^k$ (i.e., perceptual representation learning, $q(z^k | o^k)$). Second, the shared language $w$ is inferred based on the collection of all agents' internal representations $\{z^k\}_k$ (i.e., upward organization, $q(w | \{z^k\}_k)$), which, as noted below, is assumed to be achievable through communicative interactions like language games.}

\deleted{where}\added{Here,} $p(\cdot)$ and $q(\cdot)$ are assumed to be \added{true}\deleted{original} probability density functions and approximate \added{posterior} probability density functions, respectively, following the conventional notation of variational inference. 
Here, the inference of \( q\left(z^k \mid o^k\right) \) corresponds to representation learning by the $k$-th agent, and the inference of \( q\left(w \mid \{z^k\}_k \right) \) is assumed to be achieved through a language game, e.g., naming game, within a group~\citep{taniguchi2023emergent,hoang2024emergent,inukai2023recursive}. As a whole, symbol emergence by a group is performed to estimate \( q\left(w, \{z^k\}_k \mid \{o^k\}_k\right) \) in a decentralized manner.

We extend the idea of bidirectional \deleted{causation}\added{influence} in symbol emergence systems from the structural viewpoint. The bidirectional \deleted{causation}\added{influence} in symbol emergence systems can be described as follows (Figure~\ref{fig:lang-struct}):

\textbf{\deleted{downward causation}\added{Downward constraint}}:
Language shared within a society gives top-down constraints to our internal representations via distributional semantics as a prior distribution. 
Computational studies suggest that distributional semantics embedded in language facilitate agents to form internal representations with a certain structure (e.g., \citep{Mikolov2013}). This may lead to the structural alignment of agents' internal representations  \added{in the computational agent models}.
Moreover, such internal representations give top-down constraints to how the agent interprets sensory-motor information via predictive processing.

\textbf{\deleted{upward causation}\added{Upward organization}}:
Language itself is formed through language games while the meaning of each word and sentence is drifting or adapting in an evolutionary manner through time. The distribution of language is affected by agents' internal representation systems, allowing them to communicate with each other. Assuming the CPC hypothesis, the language $w$ is formed through language games, and the games play a role of decentralized Bayesian inference. As a result, the language $w$ becomes a representation of the agents' internal representations $\{z^k\}_k$ in terms of representation learning. This means language encodes the structure of internal representations in a collective manner. \added{One influential framework for understanding the structure of such internal representations and concepts, particularly their grounding in perception and relation to language, is Gärdenfors' theory of Conceptual Spaces \citep{gardenfors2000conceptual}}. Also, the internal representation system $z^k$ of the $k$-th agent reflects the structure of sensory-motor information of the $k$-th agent's \deleted{world, or its Umwelt}\added{perceptual world}.
If the structures embedded in each agent's sensory-motor information $\{o^k\}_k$ are structurally similar, this may cause \added{the structure of} agents' internal representations to be structurally similar. Additionally, it causes the variable $w$ acting as a parameter of prior distributions of $\{z^k\}_k$ to result in having distributional semantics encoding a certain type of structure shared by the distributions of $\{z^k\}_k$.

We \deleted{argue}\added{consider} that studying this bidirectional \deleted{causation}\added{influence} between language and perceptual internal representations is essential to understanding qualia structure.
Note that we do not take a position on either side of binary claims, such as  ``language is dominant in qualia structure alignment than sensory-motor experiences,''  or ``perceptual experience is more dominant than language.'' \added{This is because language itself emerges from, and is shaped by, the latent structure of sensory-motor experiences grounded in our embodied interactions with the world.}\deleted{because language itself emerges and is created on the basis of the latent structure of sensory-motor experiences originating from the world experienced based on our own embodiment.} The \deleted{main argument}\added{central theme} in this paper is this bidirectional \deleted{causation}\added{influence}.


\subsection{Levels of \deleted{qualia}\added{visual phenomenology} and language}

\begin{table}[b]
\centering
\begin{tabular}{| m{3cm} | m{8cm} | m{4cm} |}
\hline
\textbf{\deleted{Qualia Level}\added{Levels of visual phenomenology}} & \textbf{Definition} & \textbf{Examples} \\ 
\hline
High Level & 
- Recognition of objects and their relationships within a scene
- Conceptual understanding and categorization of visual inputs & 
- A bird flying in the blue sky, a person smiling, a car driving on the road \\
\hline
Intermediate Level & 
- Groupings of visual features such as colors, edges, and textures
- Formation of contours, surfaces, and basic objects from visual features & 
- Edges of a table, the color gradient of a sunset, the texture of a fabric \\
\hline
Base Level & 
- Basic visual field composed of regions and locations
- Perception of spatial relations and simple visual elements like points and regions & 
- The empty canvas of visual space, spatial arrangement of dots on a screen \\
\hline
\end{tabular}
\caption{\deleted{Qualia Levels}\added{Levels of visual phenomenology} and Examples based on ~\cite{haun2024unfathomable}}
\label{table:qualia_levels}
\end{table}

\added{The nature of the relationship between perceptual experiences and linguistic expressions varies, suggesting that the downward constraints imposed by language might differentially affect various aspects of subjective experience. While the emergence and sharing of language, potentially developed through interactive processes like language games, likely relies on and co-evolves with some form of shared representational structures enabling communication~\citep{Steels2005,taniguchi2016symbol,taniguchi2024collective}, this does not necessarily mean all aspects of experience are equally constrained. It remains plausible that more conceptual or categorized aspects of perception are significantly shaped by these linguistic constraints, whereas more basic sensory aspects might be less susceptible. Understanding this potential variation benefits from considering distinctions within subjective experience itself.}
\deleted{Considering the variety of directness or abstractness in its relationship between perceptual experiences and linguistic expressions, e.g., words or sentences, we can consider that there are several levels of qualia and internal representations from the viewpoint of the relationship between language and qualia.
In particular, considering building language through language games, it is highly probable that we, humans, need to have a basic representation that is structurally similar among people to a certain extent. In addition, language acts as an external representation system encoding agents' observations $\{o^k\}_k$ from the viewpoint of the CPC hypothesis.}

\deleted{For further discussion, let us introduce a hierarchy of qualia adapted from  haun2024unfathomable. Table describes the three-level hierarchy of qualia. Lower level qualia are expected to have less influence from the structure of language, i.e.,downward causation. Higher level qualia are expected to be more influenced by the structure of language because their categories can have greater cultural dependency and do not emerge immediately from perceptual input.}

\added{For further discussion, let us introduce a hierarchy of \added{visual phenomenology} adapted from ~\cite{haun2024unfathomable}.
Table~\ref{table:qualia_levels} describes the three-level hierarchy of \added{visual phenomenology}. \added{Base-level visual phenomenology} is expected to have less influence from the structure of language \added{(i.e., less subjected to downward constraints from language)} because it may reflect the fundamental structure of sensory input. \added{Intermediate-level visual phenomenology} might be influenced by language to some extent because categorization processes are involved. \added{Higher-level visual phenomenology}, \added{concerning the recognition of objects, scenes, and events based on multimodal information, memory, and conceptual knowledge,} is expected to be \added{most susceptible to linguistic influence (downward constraints)}. \added{This potential susceptibility could arise from factors including its greater abstractness, the cultural dependency often involved in its categorization, and the fact that it does not emerge directly from raw perceptual input.} \added{However, it is important to acknowledge the ongoing debate within philosophy concerning cognitive phenomenology -– that is, whether conceptual thought itself possesses a distinct, non-sensory phenomenal character \citep[representing diverse views in this debate, see e.g.,][]{smithies2013nature, smithies2013significance}}. \added{While our discussion touches upon conceptually-rich phenomenology, fully engaging with the cognitive phenomenology debate is beyond the scope of this paper, which primarily focuses on the structure of perceptual (specifically visual) experience and its interaction with language.}}

\deleted{In semiotics, on which the theory of symbol emergence systems is based, Peirce proposed the classification of signs, i.e., signals. The classification of qualia in Table 1 can be compared to that of signs in semiotics. According to Peirce, signs can be classified into qualisigns, sinsigns, and legisigns, corresponding to firstness, secondness, and thirdness, which is a general trichotomy in Peirce's philosophy, respectively. Qualisigns refer to a quality as a base-potentiality of being perceived as something. In this context, the base level corresponds to the qualisign. Sinsigns emerge each time a perceiver interacts with the qualities, corresponding to perceived colors or shapes in the intermediate level. Finally, legisigns are linguistic meanings or concepts that arise in relation to colors, shapes, and conventions. For example, if the perceiver who perceives colors or shapes finally regards the object as a cup, the concept of cup is a legisign. There is a hierarchical relationship in which legisigns are based on sinsigns which are, in turn, based on qualisigns.}

From the viewpoint of language understanding, we have far higher levels of internal representations, like abstract concepts. The issues of qualia about concepts are controversial within consciousness research, e.g., \cite{kemmerer2016language,mcclelland2016concepts}.
Higher representations become more abstract.
Importantly, to build abstract concepts, we need to explain them using compositional (or syntactic) language descriptions. For example, consider the question, ``What is a democratic country?'' Utilizing the compositional and syntactic nature of language, we can build abstract concepts within society by explaining them using sentences---compositions of words, some of which are perceptually grounded---and reaching consensus.
Similarly, in the realm of qualia, lower-level qualia can influence higher-level qualia structures, while the effects of language on higher-level qualia can propagate down to perception, potentially leading to alignment across different levels of conscious experience. This is also a \deleted{downward causation}\added{downward constraint} that may be occurring in one's mind at its qualia level.

This explains the perspective on the bidirectional \deleted{causation}\added{influence} between language emergence and qualia structure alignment. 
Empirical evidence based on constructive studies and psychological studies is both required to clarify this point.
In the following, we discuss the existing studies and future directions of this field.

\section{Constructive Approach}
\subsection{Overview}
In this paper, we propose to investigate the bidirectional \deleted{causation}\added{influence} between language and qualia structures based on a constructive approach. The basic research strategy is to train neural network models with certain learning mechanisms (e.g., CPC introduced in the previous sections) and to extract the internal representations of the neural network to external inputs. Then, we compare the structure of the internal representation of the trained models with the qualia structure of humans estimated by behavioral experiments. If we find a certain degree of sameness between the structure of the internal representations of the model and the qualia structure, we can say that these learning mechanisms used in the model could also be used in humans.

The goal of this constructive approach is (1) to understand which learning mechanisms can generate the internal representations that resemble human qualia structures, (2) to understand the properties of the internal structures in more detail. In the model, by systematically manipulating the learning mechanisms and the stimuli used to train and test the model, we can pursue these questions in a way that is not possible in real experiments. We should note that the goal is not to create a ``conscious AI''. Even if we create a neural network model whose internal representation mimics the human qualia structure, this does not necessarily mean that the neural network model has consciousness and that its conscious experiences are similar to those of humans. All we can say is that the structural correspondence is one of the necessary conditions, but not a sufficient condition, for a neural network model to have conscious experiences similar to those of humans. Accordingly, this paper only proposes to use neural network models as a tool to infer the mechanisms of learning. 

\begin{figure}
    \centering
    \includegraphics[width=0.8\linewidth]{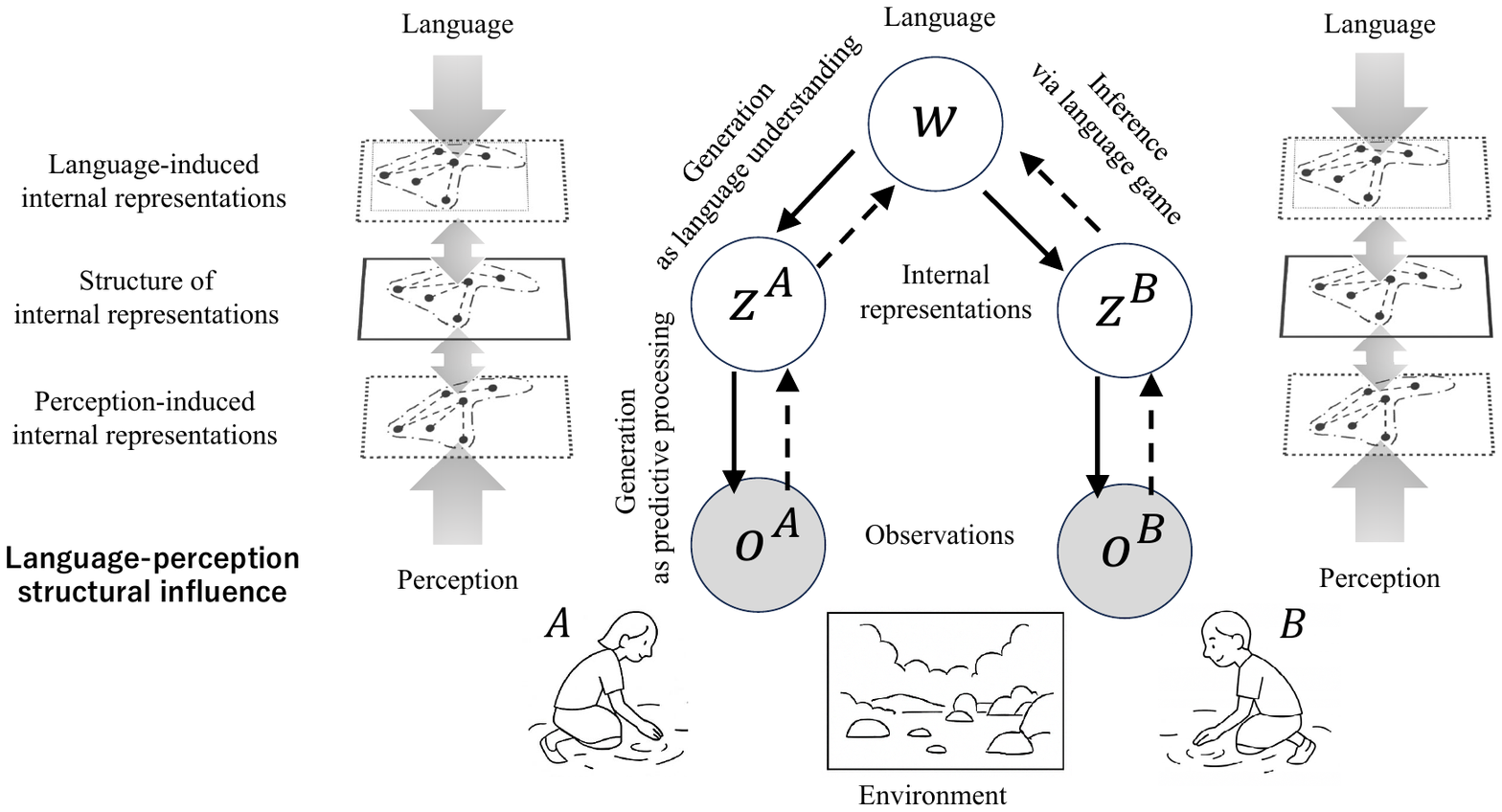}
    \caption{\deleted{\textbf{Structural influence in qualia and language}: (A) Language-perception structural influence within an individual: The bidirectional arrow represents the mutual influence between the two qualia structures induced by language and perception. (B) Inter-personal structural influence: The diagram shows how language acts as a medium for aligning internal representations (qualia structures) between individuals A and B having a shared language system as a prior. The language system is updated through \deleted{semiotic communication}\added{communication using language} over time.}\added{\textbf{Bidirectional influence between language, perception, and internal representations.} This figure illustrates how the structure of an individual's internal representations ($z$) is shaped by both bottom-up processing of perceptual observations ($o$) from the environment and top-down constraints from language (illustrated conceptually on the left and right sides). Simultaneously, the language emergence process shown in the center depicts how internal representations ($z^A, z^B$) formed by multiple agents (A, B) observing a shared environment can inform the emergence of a shared language system ($w$) through communicative interaction (e.g., language games). This emergent language system ($w$) then acts as a prior, imposing downward constraints that influence individual representations ($z^A, z^B$) and can promote their alignment across agents.}}
    \label{fig:influence}
\end{figure}

Figure~\ref{fig:influence} illustrates the structural influence in qualia and language through a probabilistic generative model representing language emergence, following the theory of CPC. The diagram depicts the language-perception structural influence within an individual. The bidirectional arrows represent the mutual influence between two qualia structures, one inferred from language and the other from perception. These structures together form an integrated qualia structure influenced by both sources.
%
The figure also  shows the inter-personal structural influence, demonstrating how language acts as a medium for aligning internal representations, and consequently, qualia structures between individuals who share a common language system. This aspect relates to how the qualia structures of different agents might be shaped and aligned through linguistic interaction. The diagram emphasizes that the language system itself is not static but is continuously updated through ongoing \deleted{semiotic communication}\added{communication using language}, reflecting the internal representation formed through interactions with the environment.

\subsection{Qualia Structure and Representational Alignment}
Comparing internal representations of humans and models has become a hot topic (see \cite{Sucholutsky2023-fx} for a review), partly due to the recent rapid development of deep learning. In addition to the developments in machine learning, construction and public-sharing of large datasets about human behaviors make human-machine comparisons feasible and meaningful. In particular, the creative use of online cloud sourcing has generated massive behavioral datasets of unprecedented size. For example, Hebart et al. created similarity judgment datasets on 1,854 natural objects by collecting about 6,000,000 similarity judgment responses \citep{Hebart2020-ft}, and Roads and Love created similarity judgment datasets on 50,000 images from the ILSVRC validation dataset by collecting about 400,000 similarity judgment responses \citep{Roads2020-ls}. As another example, Kawakita et al. created similarity judgment datasets of 93 colors from color-neurotypical and atypical participants, a total of about 700 participants \citep{kawakita2025my, Kawakita2023-xe}. Such massive datasets, in which qualia structures can be estimated, should be readily used to investigate the mutual influence between language and internal representations discussed in this article.

A standard method for comparing different representational structures is Representational Similarity Analysis (RSA) \citep{Kriegeskorte2008-ti}. RSA compares representational structures, such as representational similarity matrices for the same stimulus set, obtained as behavioral data, neural measurements, or model outputs. Then, the representational similarity matrices are compared based on, for example, a correlation coefficient. A high correlation indicates the similarity of the two representational structures. Typically, this type of comparison assumes predefined correspondences, such as a given item in one domain is correspondingly represented as the same item in the other domain. Such comparisons are called ``supervised comparison'' or ``supervised alignment''. Supervised comparison (or alignment) gives us a rough idea of the similarity between representational structures. 

To evaluate a stronger or finer level of similarity, we should consider ``unsupervised alignment'' methods \citep{kawakita2025my, sasaki2023toolbox}. Unsupervised alignment does not assume correspondences between the representations of the stimuli beforehand, but rather finds optimal correspondences during the optimization processes of the Gromov-Wasserstein distance\footnote{\added{The Gromov-Wasserstein distance is a metric used in optimal transport theory to compare the structure of two metric measure spaces without relying on any prior point-to-point correspondence, making it suitable for unsupervised comparison of relational structures \citep[see e.g.,][]{memoli2011gromov}.}}. This allows us to test the validity of the very assumption that, for example, the internal representations of the set of same stimuli should correspond to those inferred from behavioral responses. For example, Kawakita et al. explored the possibility that the experience of colors might be relationally different between different individuals \citep{kawakita2025my}. In assessing the similarity between the structures of human qualia and the internal representations of models, we encourage the use of unsupervised alignment to assess a stronger level of structural similarity or a finer level of structural difference beyond a simple correlation coefficient \citep{Kawakita2023-xe,sasaki2023toolbox,takahashi2024self}.

Here we review one of the attempts to use unsupervised alignment methods to assess whether the internal representations of neural network models resemble the representational structures of human behavior. Using the aforementioned large dataset of 93 color similarity judgments, Kawakita et al. showed that a state-of-the-art multimodal language model (GPT4) has strikingly similar representations of 93 colors to those of color-neurotypical humans, to the point where the two representations are almost perfectly aligned in an unsupervised manner \citep{Kawakita2023-xe}. On the other hand, GPT3.5, which is an earlier version and an unimodal language model without vision, did not have similar representations unlike GPT4. Yet, GPT3.5 still has better alignment performance compared to popular simple color models such as RGB and CIELAB. At this point, it is unclear which factors of GPT4 make such remarkably good alignment with humans. In the context of the topic in this paper, it is intriguing to investigate whether such internal representations can be obtained through visual learning alone (i.e., \deleted{upward causation}\added{upward organization}), or they require a combination of language and vision learning (i.e., bidirectional \deleted{causation}\added{influence}).

\subsection{\deleted{Intra- and Inter-personal structural influence}\added{Structural Influence of Language on Internal Representations}}\label{sec:influence}

As depicted in Figure~\ref{fig:influence}, we can identify two pathways of influence, through which qualia structures can be influenced: via perception or via language. Recent empirical evidence based on constructive approaches has been accumulating, as described in the following \added{discussion}\deleted{subsections}. In a way, this complements the psychological research reviewed in Section 2.2. 

\deleted{\textbf Intra-personal structural influence}

Intra-personal structural influence, or language-perception influence, has been extensively studied in machine learning and cognitive robotics. 
Note that the studies referred to in this section assume the language system itself is static and given, and do not consider the dynamics of the language system itself.

Cognitive systems can form internal representations from perceptual sensations. Neural networks and probabilistic generative models can form internal representations from multimodal data, including images, through representation learning. Variational autoencoders and disentangled representation learning are representative approaches in this domain~\citep{kingma2014stochastic,rezende2014stochastic,srivastava2017autoencoding,oord2017neural,suzuki2016joint,suzuki2022survey,higgins2016beta,mathieu2019disentangling,carbonneau2022measuring}. In the context of symbol emergence in robotics and developmental robotics, studies have shown that robots obtaining multimodal perceptual data (visual, audio, and haptic) can form object categories in a bottom-up manner~\citep{taniguchi2016symbol,nakamura_grounding,Araki12,Ando13,nakamura2012bag,MLDA+NPYLM,taniguchi2020neuro}. 
It was often reported that simultaneous learning of language and multimodal sensory data results in different internal representations in these studies \added{compared to learning from sensory data alone. This corresponds to the 'upward organization' pathway where experience shapes internal structures.}

Modeling the distribution of linguistic corpora leads internal representations to form a certain type of structure. Distributional semantics~\citep{harris1954distributional,firth1957synopsis,boleda2020distributional}, a key concept in recent natural language processing successes, has revealed that neural networks can discover syntactic-semantic structures in the latent space by modeling the distribution of word sequences in large corpora. Word2vec, for instance, can solve analogy tasks by capturing relationships between words in vector space~\citep{Mikolov2013a,Mikolov2013}. 
\cite{lample2018word} showed that structural similarity between distributional internal representations of word embeddings allows machine translation systems to perform word translation without parallel data, relying on distributional semantics of different languages.
Large language models (LLMs) are believed to exploit this nature, capturing high-level latent structures of language. Recent studies have shown that LLMs can form perceptual structures similar to humans, including color representations \citep{marjieh2023large,Loyola2023-ow,Kawakita2023-xe,Huh2024-ax}.

Multimodal LLMs, such as vision-language models, merge language and perceptual representation learning. Contrastive Language–Image Pre-training (CLIP), trained on contrastive learning between language and vision information, has proven effective in various tasks across natural language processing, computer vision, and robotics~\citep{radford2021learning}. CLIP's training approach, which increases cosine similarity between internal representations formed by language and vision, serves as a \deleted{constructive}\added{computational} model of language-perception structural influence. \added{Theoretically, frameworks like predictive coding also suggest mechanisms where language could act as a prior, providing top-down constraints that shape perceptual inference and the resulting internal representations \citep{hohwy2013predictive,friston2019free}.}

\deleted{\textbf Inter-personal structural influence}

\deleted{Inter-personal structural influence has fewer constructive approach-based evidence. cite{bouchacourt-baroni-2018-agents} studied visually-grounded referential games, showing that language emergence led to aligned internal representations among agents, although these representations did not capture conceptual properties of the input. cite{Kouwenhoven2024-dk} introduced an alignment penalty in the learning process, measured using RSA, which improved agents' performance on compositionality benchmarks. These studies were conducted in the context of emergent communication. 
They explicitly considered dynamically changing language and alignment of internal representations.
Further studies in emergent communication are crucial for investigating inter-personal structural influence with a constructive approach.}

\deleted{These studies with constructive approaches collectively demonstrate the complex interplay between language, perception, and internal representations, both within individuals and across communicating agents. They can be taken as evidence for bidirectional \deleted{causation}\added{influence} between qualia structure formation and language emergence.}

\added{A concrete example illustrating how shared symbols emerging from communication can exert downward constraints on individual representations comes from studies using probabilistic generative models and naming games. For instance, \cite{taniguchi2023emergent}, using a Metropolis-Hastings naming game (where agents converge on shared names based on joint attention, representing upward organization), demonstrated that these emergent shared signs subsequently influenced the agents' internal perceptual representations (learned via VAEs). Specifically, the communication process led to changes in the distribution of internal representations such that categorization became more consistent between the two agents, demonstrating a form of downward constraint where the emergent symbol system shapes individual representations \citep{taniguchi2023emergent}.}

\added{Reflecting on such constructive examples naturally leads to implications for inter-personal alignment. If multiple individuals within a community are similarly influenced by the downward constraints imposed by a shared, emergent linguistic system (like the signs developed in the naming game), it follows that their internal representational structures would tend towards greater coordination or similarity with each other. Indeed, computational studies in emergent communication provide supporting evidence for this possibility; for example, visually-grounded referential games have shown that language emergence can lead to aligned internal representations among agents (though challenges remain in capturing deeper conceptual properties)~\citep{bouchacourt-baroni-2018-agents}, and explicitly encouraging alignment during the learning process can improve performance on tasks requiring compositionality~\citep{Kouwenhoven2024-dk}. While these studies demonstrate this potential, the emergence of more complex compositional or syntactic language and its potential to align the deeper relational structure of internal representations across individuals remains a significant challenge and direction for future constructive research. Nonetheless, this perspective highlights the potential pathway from individual experience, through emergent communication, to inter-personally coordinated representational structures, outlining a key aspect of the bidirectional influence this paper explores.}

\cite{Huh2024-ax} proposed the Platonic Representation Hypothesis, which suggests that internal representations formed by language and vision result in similar latent structures, arguing for a pre-existing structure of the world independent of perceptual modalities and different deep neural networks converge toward having structurally similar internal representations if they are sufficiently deep and trained with a sufficient amount of data. However, when considering the bidirectional \deleted{causation}\added{influence} in symbol emergence systems, the data they presented as evidence for their hypothesis can be interpreted differently. If the CPC hypothesis holds, distributional semantics in language encode structural information about the world (see Figure~\ref{fig:birateral-cpc})~\citep{taniguchi2024collective}. Language emerges from perceptual information (i.e., \deleted{upward causation}\added{upward organization}) and subsequently influences internal representations (i.e., \deleted{downward causation}\added{downward constraint}), resulting in similar latent structures formed by language and vision. It is important to note that their hypothesis was evaluated using a vision-image dataset developed mainly for image-captioning tasks, where the similarity in latent representations is most evident when the target data is closely correlated. In this way, the Platonic Representation Hypothesis can be seen as an alternative explanation for inter-personal structural influence.

\added{In summary, constructive approaches provide valuable tools and evidence for exploring both the upward organization of internal representations from experience and the downward influence of language structure on those representations within individuals. They also offer a framework for hypothesizing about the consequent effects on representational coordination across individuals sharing an emergent language.}

\subsection{\added{Future Directions for Constructive Investigation}\deleted{Unified framework for intra- and inter-personal structural influence}}

\added{The constructive approach, as outlined in this paper, offers a powerful methodology for moving beyond correlations to investigate the potential causal mechanisms underlying the bidirectional influence between language emergence and the structure of subjective experience (qualia structure). By building and manipulating computational models, we can systematically test specific hypotheses about these complex dynamics. This section outlines key directions for future research using this approach.}

\deleted{The constructive approach can test causal hypotheses based on models that incorporate dynamic and complex causal relationships.} For \deleted{upward causation}\added{upward organization} (Figure~\ref{fig:lang-struct}~(a)), \added{a crucial direction is to develop} \deleted{it is crucial to study} \added{computational}\deleted{constructive} models that \added{explain how language structures featuring properties like distributional semantics or basic syntax can emerge from agents' interactions grounded in their perceptual experiences.} \deleted{enable the emergence of distributional semantics.} While neural network-based language models are known to capture structural relationships of word meanings in vector spaces by learning word distributions (see Section~\ref{sec:influence}), 
\deleted{it remains unclear how language emergence can create a language system with such distributional semantics.} \added{bridging the gap between interaction-based emergence and the creation of semantically rich linguistic systems remains a key challenge.}

For \deleted{downward causation}\added{downward constraint} (Figure~\ref{fig:lang-struct}~(b)), \added{constructive experiments should investigate how different types of linguistic input shape the relational structure of internal perceptual representations.} \deleted{it is important to investigate if multiple agents can learn visual representations simultaneously with learning a language model, and to examine whether the agents' internal representations become more aligned compared to conditions without language.} \added{For example, one could train multimodal models (e.g., VAEs conditioned by language models, potentially inspired by recent Vision-Language Models) with different linguistic inputs---comparing, perhaps, simple labels versus compositional descriptions, or languages with distinct categorical structures (cf.~Sec~2.2). The resulting internal representations could then be analyzed using methods like RSA or unsupervised alignment (e.g., Gromov-Wasserstein distance, Sec~3.2) and compared against human qualia structures derived from behavioral data (e.g., similarity judgments) to test specific hypotheses about language's structuring effects.} \deleted{As introduced in Section 3.3, various related studies exist, but their implications to qualia structures remain unclear at this stage.} 

Regarding bidirectional \deleted{causation}\added{influence} (Figure~\ref{fig:lang-struct}~(c)), \added{the ultimate goal is} \deleted{it is essential} to construct \added{integrated models} \deleted{a model} that incorporates \added{the co-development of perceptual representation learning, interaction-based language emergence, and the resulting downward constraints from the emergent language back onto perception, simulating the full loop depicted in Figure~\ref{fig:birateral-cpc}.}  
\deleted{both intra- and inter-personal interactions as depicted in Figure 4.} \deleted{Using a constructive approach, we need to test how language emergence influences the alignment of qualia structures (Figure~\ref{fig:influence}).}

\added{Recent developments in modeling emergent communication with deep generative models provide valuable starting points. For instance, the Metropolis-Hastings Naming Game framework demonstrates how shared word-level labels can emerge through decentralized Bayesian inference based on joint attention, and how these emergent labels subsequently exert downward constraints influencing agents' internal representations and improving categorization consistency \citep{taniguchi2023emergent}. Extensions incorporating sophisticated Vision-Language Models (VLMs) explore related dynamics in tasks like image captioning, where communication games facilitate knowledge fusion between pre-trained models \citep{matsui2025metropolis}. Relatedly, other work explores how compositional structures might emerge in similar game settings \citep{hoang2024emergent}. Building upon these approaches, future models should aim to simulate the co-emergence of more complex, potentially compositional or syntactic, language structures and rigorously evaluate their influence on the ``relational structure'' of learned perceptual representations, going beyond simple categorization alignment.}
\deleted{Generative models of emergent communication, such as Inter-GMM+VAE, have shown that even word-level naming shared through Metropolis-Hastings naming games can influence the structure of internal representations and improve the performance of perceptual stimulus modeling citep{taniguchi2023emergent,hoang2024emergent}. In these games, agents probabilistically propose names based on their inferred internal states (representations) and accept or reject names proposed by others based on their own inferred state; this process mathematically corresponds to a Metropolis-Hastings sampling algorithm operating on an implicit joint probabilistic model encompassing both agents citep{taniguchi2023emergent}. This mechanism facilitates the emergence of shared lexical conventions and leads to consistent categorization by the agents.}
\deleted{It is important to integrate such models with language models capable of handling compositionality and syntactic structures to examine the comprehensive interdependence between language emergence and qualia structure in bidirectional influence.}
\added{A key methodological challenge will be developing robust techniques to quantify and compare the ``relational structure'' of internal representations in these models with human qualia structures derived empirically.}

\added{Ultimately, this constructive research program, tightly coupled with empirical validation using behavioral and potentially neural data, holds the potential to dissect the complex interplay proposed in this paper and provide computationally grounded insights into how shared language might contribute to shaping the structure of our subjective world.}

\section{\added{Discussion}}

\added{This paper presented a perspective centered on the bidirectional influence between language emergence and qualia structure, investigated via a constructive approach. This discussion section elaborates on the value and context of this approach, addresses key challenges and limitations, and considers broader implications.}

While psychological experiments can only find correlations and rarely claim causal relationships~\citep{pearl2018book}, the constructive approach can, in principle, examine causal influences through explicit control and interventions. This makes the constructive approach particularly attractive in studying the complex dynamics between language emergence and qualia structure formation\added{, offering an alternative to purely speculative accounts by grounding hypotheses in computational mechanisms}.

\added{The issues discussed here connect deeply with long-standing problems such as the symbol grounding problem~\citep{harnad1990symbol}, questioning how symbols acquire meaning through connection to the non-symbolic world. Our approach resonates with work in cognitive linguistics emphasizing the embodiment of meaning and the grounding of abstract concepts in perceptual experience~\citep{lakoff1987woman}, while extending these ideas through dynamic modeling of emergence and bidirectional influence.}

Attention should also be paid to the varying levels of  \deleted{qualia}\added{visual phenomenology} mentioned in Section 2.4, and the degree of dependence on language at each level. \added{Investigating how the proposed bidirectional influence manifests differently across these levels is an important challenge.} \deleted{It is crucial to design and verify model-based experimental systems that can address these suggestions.} \added{Designing constructive experiments and model-based analysis techniques sensitive to these distinctions is crucial.}

In the field of semiotics research based on psychological experiments, there is a subfield called experimental semiotics~\citep{galantucci2009esreview,galantucchigarrod2011esreview}. For example, Okumura et al. have verified the validity of a symbol emergence model based on the Metropolis-Hastings naming game through behavioral experiments~\citep{okumura2023metropolishastings}. \added{Building on such approaches, experimental semiotics may offer further avenues for empirically investigating the interdependence between language emergence processes and the formation or alignment of qualia structures.
Crucially, computational models developed through the constructive approach, such as the Metropolis-Hastings naming game model mentioned above, can play a vital role in designing informative behavioral experiments and interpreting their results within a mechanistic framework, further underscoring the value of integrating constructive and empirical methods to tackle these complex questions.}

\added{Several challenges and limitations inherent to this research program must be acknowledged. One significant challenge concerns the interpretation of findings from computational models---particularly the difficulty in distinguishing qualia structures from purely linguistic structures when comparing models to human data or other AI systems. While our framework explicitly aims to model the interaction between linguistic structure and the structure of perceptual experience (cf. qualia structure), we do not insist that computational models possess genuine qualia. The scientific value lies in using these models to investigate the emergence, interaction, and potential coordination of structural patterns that are hypothesized to correlate with subjective experience in humans. This allows analysis that goes beyond linguistic structure alone, focusing on the dynamics of potentially qualia-relevant representational geometries. Further limitations include the inherent difficulty in precisely characterizing the mapping between internal representational structures and qualia structures, given the complexities of neural coding, plasticity, and degeneracy (as noted in Footnote 2).}

Considering wider categories of qualia is another future direction for studies. While the CPC hypothesis has primarily discussed the emergence of language and qualia structure based on predictive coding of sensory signals from outside the body (i.e., external world), the bidirectional \deleted{causation}\added{influence} between language emergence and qualia structure is also exemplified by emotional experiences and categories, which are grounded to interoceptive sensation (i.e., internal world). Investigations by \cite{seth2016active} into the predictive coding of interoceptive signals have elucidated the critical function of predicting and processing prediction errors in somatic sensations for the formation of emotional experiences\footnote{Complementing this perspective, \cite{barrett2017theory}'s constructivist theory of emotion posits that, in addition to interoceptive sensations, contextual information from the external environment, particularly linguistic input, plays a fundamental role in shaping emotional categories. \cite{gendron2018emotion} further expand this conceptualization, asserting that the formation of emotional categories is not limited to individual cognitive processes but is significantly influenced by interpersonal interactions and broader sociocultural and historical factors.}.
The link between emotional qualia and predictive coding enables the application of the CPC hypothesis to illuminate the bidirectional \deleted{causation}\added{influence} between emotional experiences (as qualia) and emotional categories (as linguistic constructs). \added{While the ineffability of qualia is a fundamental challenge across all sensory domains, emotional experiences present unique additional challenges for the CPC hypothesis. Emotional qualia are heavily dependent on interoceptive sensations that may have fewer established linguistic conventions and social coordination mechanisms compared to visual or auditory qualia that are more frequently referenced in everyday communication}\deleted{The examination of emotional qualia through the lens of the CPC hypothesis presents unique challenges. In contrast to visual or auditory qualia, emotional experiences are heavily dependent on interoceptive sensations that are inherently difficult to share and verify intersubjectively}\footnote{\added{This complexity is further illuminated by studies on alexithymia and the differences in emotional processing between individuals with Autism Spectrum Disorder (ASD) and neurotypical individuals, revealing diverse relationships and structures in emotional qualia across these populations.}}.
\added{To address these intricate issues, there is an increasing need to focus on constructive approaches.}
\added{The CPC framework provides important insights into how emotional qualia are constructed and shared through linguistic interactions across populations with different bodily characteristics, social affiliations, and between typical and atypical populations. In other words, the framework explains how the multimodal sensations (such as somatic, visual, and auditory) contribute via cognitive processes to the subjective experience of emotion, and how this experience in turn contributes to the emergence of emotional words (e.g., ``happy'' and ``sad'') that are shared with others.}\deleted{This approach offers a novel methodology for investigating the complex interplay between somatic sensations, cognitive processes, and the subjective experience of emotions in both typical and atypical populations.}

\added{Despite these challenges, we believe the proposed perspective and constructive methodology offer a promising avenue for integrating insights from philosophy, cognitive science, AI, and robotics to make tangible progress on understanding the deep entanglement of language, perception, and subjective experience.}

\section{Conclusion}

In this paper, we presented a novel perspective on the bidirectional \deleted{causation}\added{influence} between language emergence and qualia structure formation, and laid out a constructive approach for investigating this complex relationship. \deleted{We proposed that language, with its syntactic-semantic structures, may have emerged through the process of aligning internal representations across individuals, and that such alignment of internal representations may, in turn, facilitate more structured language.}\added{We proposed that language, with its syntactic-semantic structures, may have emerged through the coordination of internal representations shaped by experience, and that such coordination may, in turn, facilitate more structured language.} This bidirectional \deleted{causation}\added{influence} is supported by recent advances in AI, symbol emergence robotics; the collective predictive coding (CPC) hypothesis, explored within this paper, offers a specific theoretical framework aligned with these dynamics. \added{Furthermore, we suggest that the constructive methodology detailed herein is particularly well-suited for investigating the underlying mechanisms and dynamics beyond correlational findings}. We explored how computational studies, particularly those involving neural network-based language models and multimodal language models, demonstrate the formation of systematically structured internal representations that can be shared between language and perceptual information. This perspective suggests that the emergence of language serves not only as a mechanism to create a communication tool, but also as a means to allow people to realize a shared understanding of qualitative experiences, despite their private nature. \added{While significant challenges remain in bridging computational models with subjective experience and addressing the complexities discussed, the specific research directions outlined for the constructive approach offer tangible pathways forward. Ultimately, this perspective motivates a research program aimed at understanding, through integrated computational and empirical efforts, how language emerging through social interaction becomes deeply intertwined with the very structure of our subjective world.}



\section*{Funding}
This work was supported by JSPS KAKENHI Grant Numbers JP21H04904, JP23H04835, JP23H04829, and JP23H04830. 



\bibliographystyle{Frontiers-Harvard} 
\bibliography{cpc_integrated}

\end{document}